\documentclass[a4paper]{article}
\usepackage{amsmath,bm}
\usepackage{INTERSPEECH2020}
\usepackage[noadjust]{cite} 
\usepackage{makecell}
\usepackage[normalem]{ulem}
\usepackage[colorlinks,linkcolor=black,citecolor=black]{hyperref}
\usepackage{xcolor,ulem}
\title{Improving Accent Conversion with Reference Encoder and End-To-End Text-To-Speech}
\name{Wenjie Li$^1$, Benlai Tang$^2$, Xiang Yin$^2$, Yushi Zhao$^2$, Wei Li$^2$, Kang Wang$^2$, Hao Huang$^1$,\qquad \qquad Yuxuan Wang$^2$, Zejun Ma$^2$}
\address{
  $^1$School of Information Science and Engineering, Xinjiang University, Urumqi, China\\
  $^2$Bytedance AI LAB}
\email{\{liwenjie3233,hwanghao\}@gmail.com, \\
\{tangbenlai,yinxiang.stephen,zhaoyushi,liwei.speech,wangkang,wangyuxuan.11,mazejun\}@bytedance.com}

\begin{document}

\maketitle
\begin{abstract}

Accent conversion (AC) transforms a non-native speaker's accent into a native accent while maintaining the speaker's voice timbre. In this paper, we propose approaches to improving accent conversion applicability, as well as quality. First of all, we assume no reference speech is available at the conversion stage, and hence we employ an end-to-end text-to-speech system that is trained on native speech to generate native reference speech. To improve the quality and accent of the converted speech, we introduce reference encoders which make us capable of utilizing multi-source information. This is motivated by acoustic features extracted from native reference and linguistic information, which are complementary to conventional phonetic posteriorgrams (PPGs), so they can be concatenated as features to improve a baseline system based only on PPGs. Moreover, we optimize model architecture using GMM-based attention instead of windowed attention to elevate synthesized performance. Experimental results indicate when the proposed techniques are applied the integrated system significantly raises the scores of acoustic quality (30$\%$ relative increase in mean opinion score) and native accent (68$\%$ relative preference) while retaining the voice identity of the non-native speaker.
\end{abstract}

\section{Introduction}

It is well-known that the second language (L2) learning process is profoundly affected by a well-established habitual perception of phonemes and articulatory motions in the learners’ primary language (L1) \cite{ellis1989understanding}, which often cause mistakes and imprecise articulation in speech productions. Therefore, L2 learners usually speak with a non-native accent. Accent conversion \cite{huckvale2007spoken,aryal2014can,aryal2015articulatory,felps2009foreign} is designed to transform non-native speech to sound as if it’s pronounced by a native speaker. This technique can not only enhance the intelligibility of non-native speech, but also can be used to generate “golden speech” for L2 learners to mimic in computer assisted pronunciation training (CAPT) \cite{felps2009foreign}.

Conventional accent conversion methods include voice morphing \cite{huckvale2007spoken,felps2010developing,aryal2013foreign,zhao2013feedback}, frame pairing \cite{aryal2014can,zhao2018accent}, articulatory synthesis \cite{felps2012foreign,aryal2013articulatory,aryal2014accent,aryal2015reduction}. Voice morphing aims to modify non-native speech in time or frequency domain. Specifically, the authors \cite{zhao2013feedback} utilized pitch synchronous overlap and add (PSOLA) to modify the duration and pitch patterns of the accented speech. Although this method can significantly reduce the accent, the generated speech seems to be produced neither by the source speaker nor the target speaker. To address this issue, the accent frame pairing method is proposed in \cite{aryal2014can,zhao2018accent}, where authors replaced dynamic time warping (DTW) with a technique that matches the source and the target frames based on the similarity of mel frequency cepstral coefficient (MFCC) or phonetic posteriorgrams (PPGs). The authors in \cite{felps2012foreign,aryal2013articulatory,aryal2014accent,aryal2015reduction} argued that improper articulatory movements result in observed accent, therefore non-native articulatory gestures were replaced by native counterparts, which were subsequently fed into L2 probabilistic articulatory synthesizer for generating native-like acoustic features. Although these systems have achieved good accent reduction results, they usually need parallel speech or articulatory data for training conversion model. To address this issue, a recent work \cite{zhao2019foreign}, proposed to use PPGs generated by automatic speech recognition (ASR) models for representing phonetic information. This work uses non-parallel data to train a speech synthesizer. Such a method showed significant improvement in audio quality, but it relies on another native accent audio as inputs. In fact this method is part of the voice conversion (VC) task, and the length and size of the PPGs sequence have a particular impact on the speed of model training and the quality of the generated audio. 
\begin{figure}[t]
\setlength{\belowcaptionskip}{-0.63cm} 
	\centering
	\includegraphics[width=\linewidth]{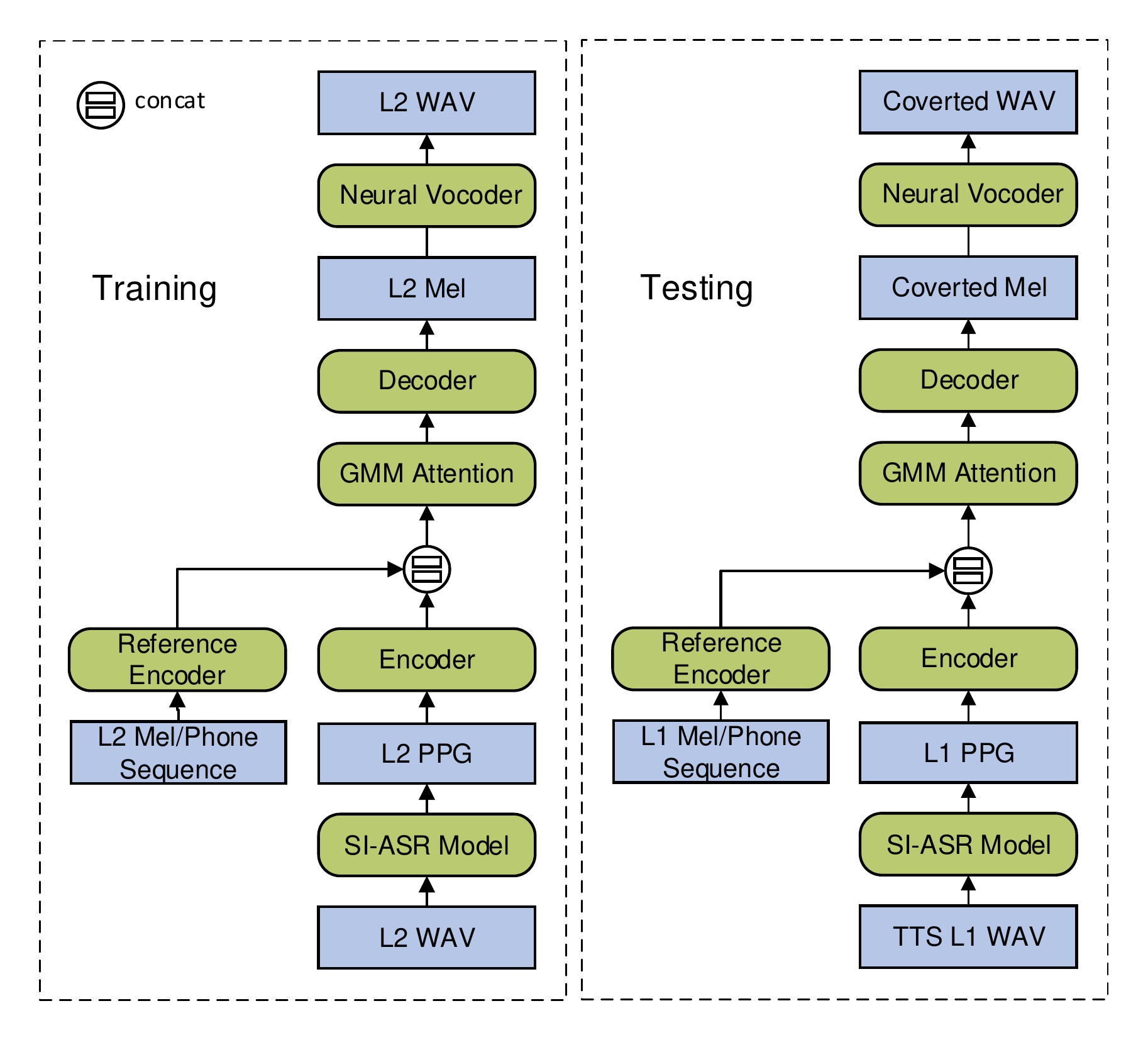}
	\caption{Overall structure of the proposed framework.
		L1: native utterance, L2: non-native utterance.}
	\label{fig:model structure}
\end{figure}
More recently, text-to-speech (TTS) decoder simultaneously receiving PPGs and speaker embedding is proposed to generate the native-accented speech without using any native utterance \cite{liu2020end}. This method needs a non-native target dependent automatic speech recognition model to generate linguistic representations. It builds a native TTS system with speaker encoder to convert non-native accent utterance to native accent. Thus, the ASR model is not robust to audio quality or noise, and mismatched TTS systems may have a severe issue with stability and similarity.

Inspired by \cite{zhao2019foreign}, we propose a robust accent conversion system by means of introducing a native TTS system and reference encoder. Since the reference text is usually available in the context of CAPT, with the help of the end to end TTS system of the native speech, the native accent speech that corresponds to the text can be easily generated. Moreover, the generated PPGs lack prosodic information, which also plays a critical role in distinguishing the accent between non-native and native speakers. We propose to use a reference encoder that is a multiple layer neural network originally applied in TTS and voice conversion \cite{skerry2018towards,lian2019towards,wang2018style}. It has been demonstrated it can effectively extract prosodic and expressive information. We apply the reference encoder module in the accent conversion to enhance the prosodic feature and intonation extraction. Additionally, to further improve performance, we optimize model architecture by using the GMM attention \cite{graves2013generating} and CBHG encoder \cite{wang2017tacotron}, which can get stable alignment and powerful feature representation. Such optimizations significantly improve the model stability and audio quality. Experimental results show that the proposed approaches are complementary and combining them to use substantially improve the quality of accent-reduced-speech.

The paper is organized as follows. Section 2 describes our proposed framework and method for accent conversion. In Section 3, experimental setup, result and analysis are displayed. The discussion and conclusion are given in Section 4.

\section{Proposed method}
The overall structure of the proposed framework is shown in Figure~\ref{fig:model structure}. Our system is mainly composed of four parts: (1) A native TTS system that generates audio from the text. (2) An ASR model that extracts PPGs from the audio. (3) A conversion model coverts PPGs into mel spectrograms. (4) A WaveRNN \cite{kalchbrenner2018efficient} vocoder models waveform from the mel spectrograms in real-time.

In the training stage, we generate the source speaker's PPGs from the ASR module. Then the conversion model maps the PPGs to the mel spectrograms. A Mean square error (MSE) loss function is used to measure the gap between the target and predicted mel. The phoneme sequences and mel features are fed into the reference encoder to get multiple source information and improve performance. In the text-available scene, during inference we can get the PPGs from the ASR model and mels from the native utterances generated by the native TTS system. Phoneme sequences and mels are fed to the reference encoder to obtain prosodic information. Finally, WaveRNN, a simple and powerful recurrent network, is used to model high-fidelity audio.

\subsection{End-To-End TTS model}
 TTS model structure is similar to Tacotron2 \cite{shen2018natural}. The TTS model is trained with the native-accented speech and corresponding text transcripts by using the mean square error (MSE) and Stop Token loss. The model incorporates phoneme features with some prosody information into an encoder LSTM and generates encoder outputs, which are then passed to a decoder LSTM with an attention mechanism to predict the mel spectrograms. A PostNet module (5 1-D convolutional layers) after the decoder is applied to predict spectral details.

\begin{figure}[t]
\setlength{\belowcaptionskip}{-0.63cm} 
\centering
\includegraphics[width=\linewidth]{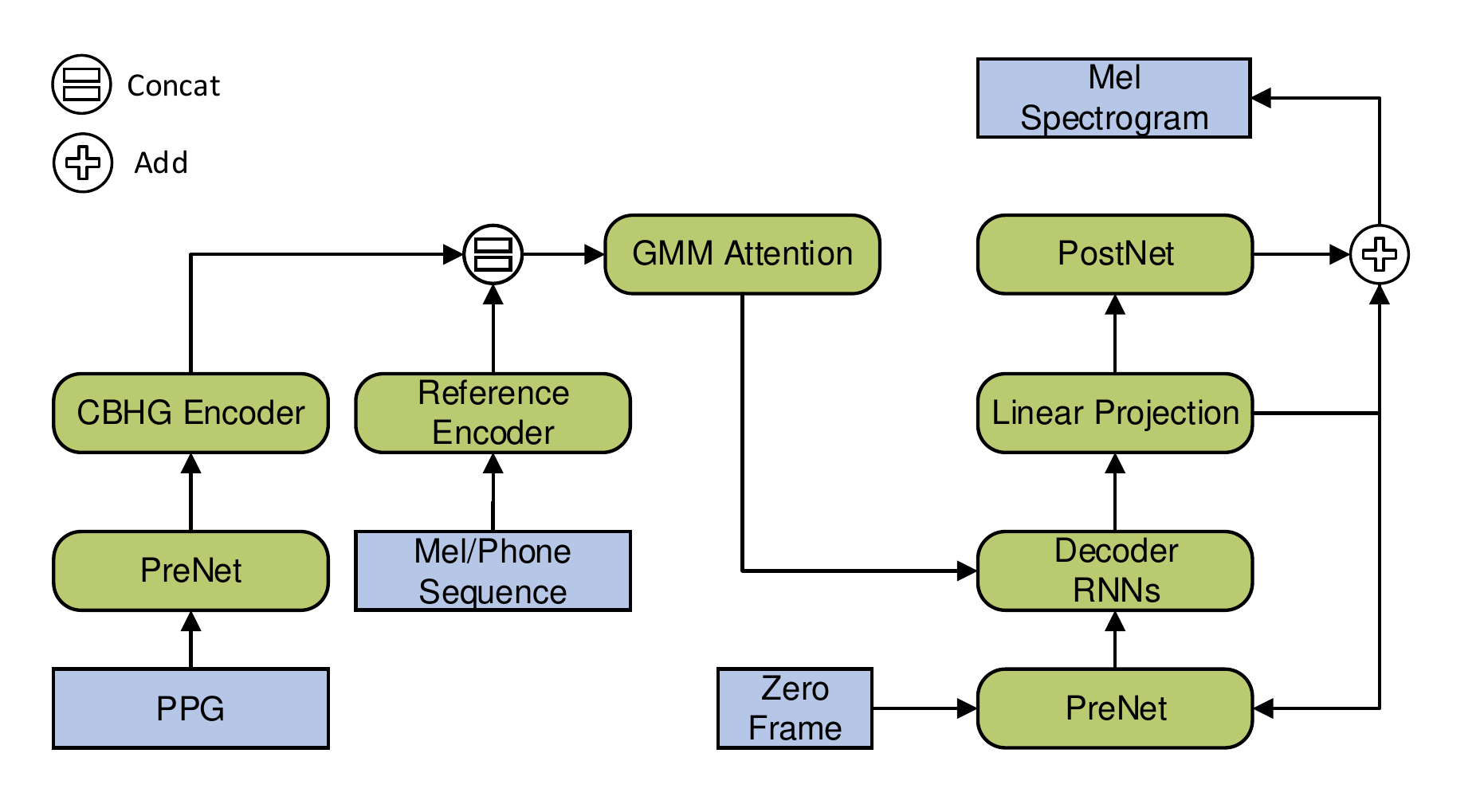}
\caption{Structure of the conversion model.}
\label{fig:conversion model}
\end{figure}
    
\subsection{SI-ASR model}  
We use speaker-independent ASR (SI-ASR) model based on DFSMN \cite{zhang2018deep} to extract PPGs from the audio. The acoustic model with objective function CTC \cite{graves2006connectionist} includes 30 layers of Deep-FSMN with 20 order look-back and 20 order look-ahead filters. The ASR model is trained on mass data from multi-accent speakers, which can generate PPGs robustly.

\subsection{Neural vocoder}
We use a WaveRNN neural vocoder to model speech waveform. WaveRNN is RNN based network capable of generating high quality speech from mel spectrograms. Compared with WaveNet \cite{oord2016wavenet}, WaveRNN has fewer parameters and can achieve real-time inference in CPU, and the quality can reach a similar level. In this case, we train the sparse WaveRNN condition on mel spectrograms. The details of WaveRNN vocoder can be referred to \cite{kalchbrenner2018efficient}.

\subsection{End-To-End conversion model}
The conversion model is illustrated in Figure~\ref{fig:conversion model}. Unlike Tacotron2 that takes character or phoneme sequences as inputs, we use PPGs sequences as inputs for the conversion model.

\subsubsection{Model modifications}
The length of a PPGs sequence is normally a few hundreds frames, significantly longer than the length of characters or phoneme sequences. Local sensitive attention (LSA) mechanism \cite{chorowski2015attention} would be unstable for such long input and has difficulties obtaining the correct alignment. As a result, the inference would generate ill-intelligible speech. One solution to this issue is to use shorter segments, for example, using word segments instead of sentences or to add a locality constraint window to the attention mechanism \cite{zhao2019foreign}. However, using word segment needs force alignments to find the word boundary in the training stage, and may lead to model failure in the inference stage, since the input is much longer than that in the training stage. For locality constraint window, in every decode time step, only part of encoder states would be used to compute attention weights. Decoder can not learn the global information of the entire encoder states, which will lead to mispronunciation. We find the problem will be more serious when the mel spectrograms and the PPGs are not equal in length.

Based on this fact, we replace the origin LSA with GMM attention. In \cite{battenberg2019location}, authors show GMM attention can generalize very long utterances. GMM attention is a purely location-based mechanism that uses mixture of $K$ Gaussians to produce the attention weights, \(\bm{\alpha_i}\), for each encoder state. The general form of this attention is shown in (\ref{eq1})
\begin{equation}
    \alpha_{i,j} = \sum_{k=1}^K{w_{i,k}\exp{(-\frac{(j-\mu_{i,k})^2}{2(\sigma_{i,k})^2}})}
    \label{eq1}
\end{equation}
\vspace{-5pt}
\begin{equation}
(\bm{\hat{\omega}_i},\bm{\hat{\Delta}_i},\bm{\hat{\sigma}_i})= Vtanh(W\bm{s_i}+b)
\label{eq2}
\end{equation}
\vspace{-10pt}
\begin{equation}
\bm{\omega_i}= \exp{(\bm{\hat{\omega}_i})}\\
\label{eq3}
\end{equation}
\vspace{-8pt}
\begin{equation}
\bm{\Delta_i}= \exp{(\bm{\hat{\Delta}_i})}\\
\label{eq4}
\end{equation}
\vspace{-8pt}
\begin{equation}
\bm{\sigma_i}= \sqrt{\exp({-\bm{\hat{\sigma}_i}})/2}\\
\label{eq5}
\end{equation}
\vspace{-8pt}
\begin{equation}
    \bm{\mu_i}=\bm{\mu_{i-1}}+\bm{\Delta_i}
    \label{eq6}
\end{equation}

  Where \(\bm{s_i}\) is hidden state of attention RNN. Intermediate parameters (\(\bm{\hat{\omega}_i}\), \(\bm{\hat{\Delta}_i}\), \(\bm{\hat{\sigma}_i}\)) are first computed using MLP in (\ref{eq2}), and then converted to final parameters using (\ref{eq3}) - (\ref{eq5}). The mean of each Gaussian component is computed using the recurrence relation in (\ref{eq6}), which makes the mechanism location-relative, and potentially monotonic when \(\bm{\Delta_i}\) is constrained to be positive. 
  
In addition to the above modification, we use the powerful CBHG encoder in \cite{wang2017tacotron} instead of original Tacotron2 encoder to extract robust linguistic representations. We found that this CBHG encoder not only makes fewer mispronunciations, but also improves the speech quality than Tacotron2 encoder.

\subsubsection{Reference encoder}
We use reference encoders to enhance the control of stress and intonation. The reference encoder stacks a 6-layer 2D convolutions with batch normalization and a bidirectional GRU layer. The number of filters in each convolutional layer is [32, 32, 64, 64, 128, 128]. We use mel and phoneme sequences as inputs. 

For mel spectrograms, we use the hidden states of the GRU as a variable-length embedding. Since it contains not only acoustic information, but also speaker identity information, we take the following measures to avoid affecting the voice timbre during conversion. First, we try to isolate the embedding from the synthesized speech. Instead of combining the embedding with the decoder, we concatenate the embedding with the output of the encoder. Second, since mel spectrograms are a compact feature, the dimension of the embedding is compressed to a minimal value. We tried four dimensions of the embedding: 1 unit, 2 units, 3 units and 4 units. Evidence shows that 4 units make better control in stress and intonation while maintaining the voice timbre of the source speaker.

Phoneme sequences are a rich source of the phonemic information to enhance the model‘s stability and improve voice quality. We compress phoneme sequences as a fixed 128-unit embedding before concatenating it with the encoder output at each time step.

\section{Experiments and results}
\subsection{Experimental setup}
The ASR model used to extract the 87-dimensional PPGs is trained on a 10,000-hour multi-speaker mixed-accent English corpus. To speed up the training and inference \cite{sak2015fast}, we stack 8 consecutive feature frames and skip 3 frames to reduce frame rate, which results in more accurate models. A 26-channel fbank with 2 pitch parameters computed using 25ms window width and 10ms shift, as well as their delta and delta-delta features are used as input features. Our conversion model is first trained on a one-hour corpus of American English to make the model more stable and for better learning of the stress and intonation patterns. The model is then fine-tuned by a female Arabic speaker of L2 English from the publicly-available L2-ARCTIC corpus \cite{zhao2018l2}. The alignment of the attention is more flexible by training conversion model using sequences of unequal lengths between the PPGs and mel spectrograms. The WaveRNN vocoder is first trained on the same American English corpus and then fine-tuned on the ZHAA dataset. The native TTS system is first trained on multi-speaker corpora before using the native corpus for fine-tuning. This L2 speaker ZHAA records 1132 English sentences out of which the first 1000 utterances are used for our conversion model and WaveRNN training. The next 50 utterances are used for validation, and the remainder is for testing. The sampling rate of all audio files is set to 16 KHz. As for the mel spectrograms, we use an 80-dimensional vector with a 10ms shift and a 50ms window. All the mel features are normalized within the range of $[-4,4]$. 
\vspace{-10pt}
\begin{table}[ht]\scriptsize
\setlength{\abovecaptionskip}{0.0 cm}
\setlength{\belowcaptionskip}{0.0 cm}
  \caption{The model details of the conversion model}
  \label{tab:conversion model}
  \centering
  \linespread{1.2}\selectfont
  \begin{tabular}{c|c}
  \toprule
  \textbf{Module}      & \textbf{Parameters}\\
  \midrule
  PPG PreNet               & \makecell[l]{2 Fully connected (FC) layers, \\128 ReLU units; 0.5 dropout rate} \\
  \hline
  Encoder CBHG             & \makecell[l]{Conv1D bank: K=16, conv-k-128-ReLU;
                                        \\Max pooling: stride=1, width=2;
                                        \\Conv1D projections: conv-3-128-ReLU,                                        \\conv-3-128-Linear;
                                        \\Highway net: 4  FC-128-ReLU, 4 FC-128-Sigmoid;
                                        \\Bidirectional GRU (128 cells)} \\
  \hline
  Mel Ref Encoder              & \makecell[l]{5 Conv2D-ReLU, 3x3 kernel, 1x2 stride;
                                        \\1 Conv2D-ReLU, 3x3 kernel, 3x2 stride;
                                        \\Bidirectional GRU (4 cells), Tanh activation} \\
  \hline
  Phone Ref Encoder              & \makecell[l]{6 Conv2D, 3x3 kernel, 1x2 stride;
                                        \\Bidirectional GRU (128 cells), Tanh activation} \\
  \hline
  Decoder PreNet           &\makecell[l]{2 FC-300-ReLU; 0.5 dropout rate}\\
  \hline
  Attention LSTM           &\makecell[l]{1-layer LSTM (300 cells); 0.1 dropout rate}\\
  \hline
  Attention            &\makecell[l]{GMM Attention, num 
gaussian=10}\\
  \hline
  Decoder LSTM           &\makecell[l]{1-layer LSTM (300 cells); 0.1 dropout rate}\\
  \hline
  PostNet            &\makecell[l]{5 Conv1D layers, 512 channels, \ kernel size=5}\\
  \hline
  Reduction factor (r) &\makecell[l]{2}\\
 \bottomrule
\end{tabular}
\end{table}
\vspace{-1pt}

Our baseline is in line with \cite{zhao2019foreign}, an accent conversion model comparable to Tacotron2. As described in \cite{zhao2019foreign}, the authors adopt an attention mechanism with a locality-constraint window size of 20 to learn the alignments between long sequences. Since the PPGs and mel spectrogram sequences are no longer equal in length, time shift of the attention window is determined by the length ratio of the PPGs and mel spectrograms. The network structure of the baseline is identical to \cite{zhao2019foreign}, apart from changing the output dimension of PPG PreNet to 128.

The structure of our proposed model is described in Table 1. Four systems will be evaluated in the experiments. In addition to the baseline system, we experiment with two ablation systems to verify the effectiveness of our work. Both the baseline and our model are trained with the Adam optimizer \cite{kingma2014adam} using a batch size of 64, configured with an initial learning rate of 0.001 and same decay strategy as \cite{vaswani2017attention}.

\begin{itemize}
\item System 1: Only structural adjustments is made. We replace the LSA with the GMM attention, and replace the original Tacotron2 encoder with CBHG network.

\item System 2: In addition to structural adjustments, the mel reference encoder is added. The structure of the mel reference encoder is described in Table \ref{tab:conversion model}.

\item System 3: In addition to structural adjustments, both the mel and phoneme reference encoders are added. The structure of the phoneme reference encoder in Table \ref{tab:conversion model}.

\end{itemize}

\subsection{Results}
We conduct three subjective listening tests to compare the performance of the systems: (1) Mean Opinion Score (MOS) test of audio quality and naturalness; (2) Voice similarity test; (3) Accent similarity test. We selected the same 25 sentences from the test set for each experiment\footnote{Audios can be found in “https://kal009l.github.io/ac-demo/”}. All raters are native speakers of English from North America. 

Audio quality and naturalness are rated on a five-point scale (1-bad, 2-poor, 3-fair, 4-good, 5-excellent) in the MOS test, aiming to evaluate the clarity and human-likeness of the speech samples. In this experiment, the participants also hear the TTS inputs and L2-ARCTIC recordings as references. Each speech sample from all systems receives 20 ratings. The results are displayed in Table \ref{tab:Mos} and Table \ref{tab:Mos_ori} with 95\% confidence interval.
\vspace{-2pt}
\begin{table}[ht]\scriptsize
\setlength{\abovecaptionskip}{0.0 cm}
\setlength{\belowcaptionskip}{0.0 cm}
  \caption{MOS results of proposed methods and baseline}
  \label{tab:Mos}
  \centering
  \linespread{1.2}\selectfont
  \begin{tabular}{c|c|c|c}
  \toprule
  \footnotesize{Baseline}   &\footnotesize{System 1} &\footnotesize{System 2} &\footnotesize{System 3}\\
  \hline
  \makecell{2.72$\pm$0.13} &\makecell{\textbf{3.52}$\pm$0.15}&\makecell{3.41$\pm$0.19}&\makecell{3.48$\pm$0.21}\\
 \hline
 \bottomrule
\end{tabular}
\end{table}
\vspace{-20pt}
\begin{table}[ht]\scriptsize
\setlength{\abovecaptionskip}{0.0 cm}
\setlength{\belowcaptionskip}{0.0 cm}
  \caption{MOS results of TTS INPUT and ZHAA recording}
  \label{tab:Mos_ori}
  \centering
  \linespread{1.2}\selectfont
  \begin{tabular}{c|c}
  \toprule
  \footnotesize{TTS INPUT} &\footnotesize{ZHAA}\\
  \hline
 \makecell{3.57$\pm$0.12}&\makecell{3.90$\pm$0.14}\\
 \hline
 \bottomrule
\end{tabular}
\end{table}

\vspace{-5pt}

The result shows that the original recordings of ZHAA and the TTS system receive higher scores. Among the conversion models, System 1 has the highest score overall whereas the baseline gets the lowest MOS rating. Compared with the baseline, the high ranking of System 1 might indicate that the system benefits from the aforementioned network structure modification. By incorporating the mel reference encoder into the conversion system, the quality of System 2 and 3 might be hurt by the unnatural intonation inherited from the TTS system which is trained only on one hour of native speech. In fact, because the mel spectrograms are used as a reference, System 2 and 3 obtain more acoustic information than System 1. As a result, System 2 and 3 are better than System 1 for audio quality, but the unnaturalness lowers the MOS scores of System 2 and 3. It is also shown that System 3 with the additional phoneme reference encoder is superior to System 2 with respect to the MOS score, suggesting that phoneme sequences have a positive effect on audio quality and prosody. 
\vspace{0pt}
\begin{table}[ht]\scriptsize
\setlength{\abovecaptionskip}{0.0 cm}
\setlength{\belowcaptionskip}{0.0 cm}

\caption{Voice similarity test results}
	\label{tab:voice similarity}
	\centering
	\linespread{1.3}\selectfont
	\begin{tabular}{c|c|c|c|c}
		\toprule
		\footnotesize{System} &\footnotesize{Baseline}   &\footnotesize{System 1} &\footnotesize{System 2} &\footnotesize{System 3}\\
		\hline
		Preference &\makecell{18$\%$} &\makecell{26.6$\%$}&\makecell{\textbf{28$\%$}}&\makecell{27.4$\%$}\\
		\hline
		Confidence &\makecell{3.93$\pm$0.19} &\makecell{4.04$\pm$0.20}&\makecell{\textbf{4.42}$\pm$0.20}&\makecell{4.23$\pm$0.21}\\
		\hline
		\bottomrule
	\end{tabular}
\end{table}
\vspace{0pt}

Table \ref{tab:voice similarity} illustrates the result of the voice similarity test. In this experiment, the participants were instructed to answer 25 questions, each of which is consisted of 5 speech samples: the original ZHAA utterance and 4 converted utterances from each system. 20 participants had to choose which converted utterance sounds more like the original recording of ZHAA. Listeners were also asked to rate their confidence level on a 7-point scale (1-not at all confident, 7-extremely confident). They were instructed to only focus on voice similarity while ignoring other irrelevant information such as accent and intonation. The result summarized in the table shows that System 1, 2, 3 have roughly the same number of votes. The acoustic information obtained by System 2 and 3 is basically the same, which converges their scores.

Table \ref{tab:accent similarity} presents the results of the accent similarity test. The participants of this experiment were asked to rate 25 sets of speech samples. In each set, the participants heard 6 speech samples: the TTS input as the reference accent, the original ZHAA recording, and the converted speech from the four systems. The listeners were instructed to rate the converted samples in terms of accent similarity with respect to the TTS input on a 5-point scale (1-very different, 5-very similar). 
\vspace{-7pt}
\begin{table}[ht]\scriptsize
\setlength{\abovecaptionskip}{0.0 cm}
\setlength{\belowcaptionskip}{0.0 cm}
  \caption{Accent similarity test results}
  \label{tab:accent similarity}
  \centering
  \linespread{1.3}\selectfont
  \begin{tabular}{c|c|c|c|c}
  \toprule
  \footnotesize{ZHAA} &\footnotesize{Baseline} &\footnotesize{System 1} &\footnotesize{System 2} &\footnotesize{System 3}\\
  \hline
  \makecell{1.81$\pm$0.20} & \makecell{2.13$\pm$0.14}  &\makecell{2.66$\pm $0.11}&\makecell{3.57$\pm$0.12}&\makecell{  \textbf{3.58}$\pm$0.10}\\
 \hline
 \bottomrule
\end{tabular}
\end{table}
\vspace{-7pt}

It can be observed from the result that the original ZHAA recording receives the lowest rating, and systems with reference encoder are superior to the system that only uses PPGs. Intonation and stress can obviously impact the listener's judgment of accent. Note that the ratings of System 2 and 3 are not much different, because phoneme sequences may not show any effect on the precise control of intonation and stress. Heavy accent and poor audio quality in the baseline system negatively affect its performance, particularly mispronunciations which may have contributed to the low accent ratings. Besides, we believe that putting all systems together for comparison makes the gap between different systems more obvious.

\section{Conclusion and future work}

In this paper, we reported an end-to-end accent conversion framework with a series of improved approaches. First to widen the applicability of the accent conversion, we employed an end-to-end TTS system generating native reference speech, which frees us from collecting speech data. To improve the quality and accent, we applied reference encoders which are complementary to PPGs, to enhance the prosodic feature and intonation extraction. After that, we attempted to modify the conversion models, such as replacing the LSA with GMM attention, and replacing original Tacotron2 encoder with GBHG encoder, to generate more stable and expressive converted speech. To fully evaluate our proposed method, we conducted experiments with 3 modified systems. All the 3 systems achieved obvious improvement in listening tests. Our System 1 mainly improved audio quality. In addition to the overall amelioration in quality, System 2 and System 3 improved in voice similarity and accentedness simultaneously. More importantly, our models can not only convert non-native speech into native speech, but also mimic the native accent.

More efforts need to be put into our work to expand the application of this technology. For example, the limited resource of the non-native target accent data is a challenge we are faced with in some cases. Second, the loss of voice similarity still exists and has a significant impact on testing. In future, some adaptive training or voice cloning methods can be used to deal with the scarce resources of the non-native data. In response to the lack of similarity in the converted voice, speaker loss and variational autoencoder can be applied to further separate speaker information from the mel spectrograms.

\bibliographystyle{IEEEtran}

\bibliography{bib1}


\end{document}